\documentclass[lettersize,journal]{IEEEtran}
\usepackage[T1]{fontenc}
\usepackage{graphicx}
\usepackage{subcaption}
\usepackage{color}
\usepackage{textcomp}
\usepackage{enumitem}
\usepackage{csquotes}
\usepackage{hyperref}

\usepackage{caption} 
\usepackage{tabularx}
\usepackage{float}
\usepackage{array}  
\usepackage{booktabs} 
\usepackage{multirow}
\usepackage{makecell}
\usepackage{arydshln} 
\usepackage{todonotes}
\usepackage{eurosym}
\usepackage{balance} 

\usepackage{siunitx}
\usepackage{cite}
\begin{document}


\title{Touch and Tell: Multimodal Decoding of Human Emotions and Social Gestures for Robots}

\author{Qiaoqiao Ren$^{1}$\and,
Remko Proesmans$^{1}$\and,
Yuanbo Hou$^{2}$\and,
Francis wyffels$^{1}$\and, and
Tony Belpaeme$^{1}$

\thanks{Faculty of Engineering and Architecture, $^{1}$IDLab-AIRO, Ghent University -- imec, Technologiepark 126, 9052 Gent, Belgium, $^{2}$Department of Engineering Science, University of Oxford}

\thanks{The authors acknowledge the use of generative AI in preparing this manuscript. Specifically, Grammarly and GPT-4o were used to assist with grammar checking and enhancing the overall readability. All content was subsequently reviewed and edited by the authors, who accept full responsibility for the final version of the manuscript.}

}

\maketitle

\begin{abstract}
Human emotions are complex and can be conveyed through nuanced touch gestures. Previous research has primarily focused on how humans recognize emotions through touch or on identifying key features of emotional expression for robots. However, there is a gap in understanding how reliably these emotions and gestures can be communicated to robots via touch and interpreted using data-driven methods. This study investigates the consistency and distinguishability of emotional and gestural expressions through touch and sound. 
To this end, we integrated a custom piezoresistive pressure sensor as well as a microphone on a social robot. Twenty-eight participants first conveyed ten different emotions to the robot using spontaneous touch gestures, then they performed six predefined social touch gestures. Our findings reveal statistically significant consistency in both emotion and gesture expression among participants. However, some emotions exhibited low intraclass correlation values, and certain emotions with similar levels of arousal or valence did not show significant differences in their conveyance. To investigate emotion and social gesture decoding 
within affective human-robot tactile interaction, we developed single-modality models and multimodal models integrating tactile and auditory features. A support vector machine (SVM) model trained on multimodal features achieved the highest accuracy for classifying ten emotions, reaching 40\,\%.
For gesture classification, a Convolutional Neural Network - Long Short-Term Memory Network (CNN-LSTM) achieved 90.74\,\% accuracy. Our results demonstrate that even though the unimodal models have the potential to decode emotions and touch gestures, the multimodal integration of touch and sound significantly outperforms unimodal approaches, enhancing the decoding of both emotions and gestures.

\end{abstract}


\section{Introduction}

\IEEEPARstart{U}{n}derstanding human emotions is a fundamental goal in the development of socially interactive robots. As robots become increasingly integrated into domains such as healthcare, therapy, customer service, and companionship, their ability to perceive and appropriately respond to human affect is becoming essential. Affective computing, which focuses on systems that recognize and process emotional signals, plays a critical role in enabling more natural and compelling human-robot interaction~\cite{gervasi2023applications}. Emotion recognition has been widely explored through modalities such as facial expressions, speech, and body gestures~\cite{noroozi2018survey}. Previous research has shown that intelligent virtual agents can be endowed with empathic behaviours to serve in roles such as virtual therapists~\cite{ranjbartabar2019first}. However, affect recognition through touch remains relatively underexplored, despite its central importance in human social communication.


Touch serves as a powerful and nuanced channel for emotional and social exchange. It plays a vital role in forming interpersonal bonds, navigating social roles, offering reassurance, and conveying empathy or urgency~\cite{hertenstein2006touch}. The communicative power of touch lies in its parameters—pressure, duration, frequency, and location—which together shape the emotional message being conveyed~\cite{hertenstein2006touch, van2015social}. For instance, a brief pat may express comfort, whereas a firm, prolonged grip may signal support or solidarity.

Recent studies have increasingly highlighted the emotional and communicative value of touch in interactive systems. For example, emotional expressions of virtual agents have been shown to influence how humans physically interact with them in VR settings~\cite{ahmed2020touching}, and tele-touch devices have enabled the remote transmission of affect~\cite{cabibihan2015physiological}. Affective cues have also been inferred from everyday smartphone touch gestures, such as taps and swipes, to assess stress levels~\cite{ciman2016individuals}. Furthermore, agents combining touch with facial expressions and speech can convey empathy more effectively, though in some contexts, touch alone may be the most influential cue~\cite{bickmore2010empathic}. Experimental research using robotic arms has shown that specific touch features (e.g., force, speed, amplitude) shape perceptions of arousal and valence, and can even override conflicting emotional cues from other modalities~\cite{teyssier2020conveying}. Physiological studies further reveal that social touch elicits distinct autonomic responses compared to self- or object-touch, indicating its unique role in emotional communication~\cite{candia2025autonomic}. Moreover, affective touch delivered through a robot has been shown to reduce users’ heart and respiration rates and increase their sense of calm and happiness, reinforcing its therapeutic potential~\cite{sefidgar2015design}.

In addition to emotional expression, researchers have studied the classification of social touch gestures, recognizing their capacity to communicate intent, build intimacy, and support remote interaction~\cite{gallace2010science}. People can express meaning through structured haptic codes~\cite{eid2015affective}, and touch-based gesture decoding has emerged as a growing field. The Social Touch Gesture Challenge 2015, for instance, provided annotated tactile data on various social gestures performed on pressure-sensitive surfaces of different textures and shapes~\cite{ta2015grenoble}. This lays the groundwork for building systems capable of decoding the social and emotional meaning behind touch, a necessary step toward enabling robots to respond to human touch in a contextually appropriate and socially meaningful way.

These findings collectively underscore the importance of decoding the emotional and social meaning of touch gestures, both as an independent modality and as part of multimodal interaction, to develop socially aware and emotionally intelligent robots.

\subsection{Robot touch sensing capabilities}

Advancements in robotics and sensor technology have enabled more complex human-robot interactions. These sensors help us build robots with tactile sensitivity, allowing them to detect and respond to human emotions through touch. Tactile sensors capture nuances like pressure and texture, vital for emotion recognition~\cite{yohanan2012haptic}, while auditory feedback from touch gestures enhances emotion decoding accuracy~\cite{ju2021haptic}. One approach for creating robot touch sensing capabilities involves covering a robot with artificial skin that simulates the human touch sensory system\cite{dahiya2009tactile}. Alternatively, robots can be embedded with sensors to measure temperature, proximity, and pressure\cite{huisman2013tasst}. 

Previous research designed soft tactile sensors, which are equipped on the iCub robot and successfully used for shape exploration~\cite{tomo2018new, schmitz2011methods}. Naya \textit{et al.} built an artificial skin by covering the robot with a gridded pressure-sensitive ink sheet~\cite{naya1999recognizing}. Using features such as absolute pressure and temporal pressure differences, they distinguished five touch gestures: \enquote{Pat}, \enquote{Scratch}, \enquote{Slap}, \enquote{Stroke}, and \enquote{Tickle}. Silvera \textit{et al.} further developed the concept of artificial skin using electrical impedance tomography, which allowed larger areas of a robot to be covered and enabled the extraction of information such as location, duration, displacement, and intensity of touch~\cite{tawil2011touch}. In addition to academic developments in tactile sensing, the industry has also explored ways to equip robots with sensing capabilities. For example, solutions like AIRSKIN~\cite{airskin2024} and Dobot SafeSkin~\cite{dobotsafeskin2024} aim to enable safe and responsive physical interaction.

Simulating touch perception capabilities using sensors has already been extensively applied in social robotics. For example, the Huggable is a robotic companion designed in the form of a teddy bear, equipped with a combination of temperature, electric field, and force sensors to facilitate affective haptic communication between individuals~\cite{stiehl2005design}. The Haptic Creature is a furry, lap-sized social robot embedded with pressure sensors throughout its body, enabling it to communicate with the world through touch~\cite{yohanan2012role}. Paro, a social robot modelled after a seal, is equipped with sensors that allow it to respond to touch and distinguish between three different gestures: petting, stroking, and hitting~\cite{shibata1996emotional}.

\subsection{Related work}

\subsubsection{Conveying emotions}

Several studies have explored how distinct emotions can be conveyed through various touch gestures~\cite{ren2023low, hertenstein2009communication, teyssier2020conveying}. For example, Yohanan and MacLean developed the Haptic Creature to detect touch and movement, investigating which gestures from a predefined set participants believed were appropriate for expressing nine specific emotions~\cite{yohanan2012haptic}. 
Some studies instructed participants to express specific emotions through touch while recording the touch sounds, from which vibrotactile feedback patterns were derived~\cite{ju2021haptic}. The aim was to determine if subjective feelings could be communicated via these simple vibrotactile patterns: the average accuracy across four emotions was 57.74\,\% by a human decoder.
In other work, couples were asked to perform gestures conveying six emotions to one another. The couples' recognition accuracy was 57\,\%~\cite{hauser2019uncovering}. However, some researchers point out that direct touch as a mode of communication is influenced by context; research indicates that haptic behaviours by themselves are insufficient for recognizing specific emotions~\cite{olugbade2023touch}. In one notable study, researchers catalogued 23 distinct types of tactile interactions, including hugging, squeezing, and shaking~\cite{hertenstein2009communication}. The findings from this study revealed that tactile behaviours on their own do not effectively distinguish between different emotions. For instance, the act of stroking was associated with expressions of sadness, as well as love or sympathy. The study also highlighted that factors such as the duration and intensity of the tactile interaction need to be adjusted to better differentiate between emotions.


\subsubsection{Conveying social gestures}

Some researchers have investigated the conveyance of gestures to a mannequin or a robot~\cite{silvera2014interpretation, yohanan2012haptic}. In previous studies, an artificial skin was developed to cover the arm of a mannequin, and the Logit-Boost algorithm was used to decode eight social touch gestures with an accuracy of 72\,\%, compared to human recognition accuracy of 90\,\%~\cite{silvera2014interpretation}. Additionally, the Social Touch Gesture Challenge 2015 was organized, where the highest accuracy achieved in the competition was 70.9\,\% for recognizing eight-second-long gestures, covering six touch gestures plus a \enquote{no touch} condition~\cite{ta2015grenoble}. This was further explored in~\cite{ren2023low}, a classification accuracy of 72.8\,\% was achieved in 2.67 seconds. Furthermore, the Haptic Creature~\cite{yohanan2012haptic} was able to distinguish between four types of touch gestures—\enquote{pat}, \enquote{Stroke}, \enquote{Poke}, and \enquote{Squeeze}—with an average accuracy of 77\,\%.

Earlier research identified specific tactile behaviours associated with each emotion through extensive behavioural coding~\cite{hertenstein2006touch}. For instance, sympathy was linked to stroking and patting, anger to hitting and squeezing, disgust to a pushing motion, and fear to trembling~\cite{hertenstein2009communication}. Additionally, previous studies have explored how individuals use tactile hand gestures to communicate emotions, providing evidence that supports the connection between emotions and gesture features such as intensity, temporal frequency, and spatial frequency~\cite{joung2011tactile}. This research aimed to investigate the usability of tactile hand gestures for emotional online communication.

\subsection{Research goal}

Previous studies primarily focused on either tactile or auditory channels in isolation and examined the correlations between social touch gestures and emotions. 
In this work, we explore the consistency of expressions of emotion based on touch and sound features using multimodal methods.
We also concentrate on decoding different emotions without providing specific social gesture options, allowing participants to express different emotions without being restricted to certain gestures.

We focus on the following research questions (RQs): 

\begin{enumerate}

    \item  Can we create touch-sensing capabilities for the Pepper robot?

    \item Are different people consistent in the way they express emotions and gestures through touch? 
    
    \item Which emotions and gestures are distinguishable based on touch? Which ones are easier to decode? 
    
    
    \item Can multimodal models be used to decode different emotions and gestures? Is a multimodal model better than a unimodal model?

\end{enumerate}

Section~\ref{sec2:Materials} outlines the experimental design for human-robot interaction, including research materials (RQ1), experimental procedures, as well as data collection and processing methods. 
In Section~\ref{data_analysis}, the collected data is analysed.
We examine to what extent emotions and gestures are expressed consistently by multiple participants (RQ2), as well as the participants' subjective notions of correlations between emotions and gestures. Afterwards, Section~\ref{s:decoding_emotions} presents the decoding of emotions and gestures using unimodal and multimodal data-driven methods (RQ3, RQ4). 
Section~\ref{sec:discussion} discusses our findings, reflects on the limitations of our research, and the paper concludes with Section~\ref{sec:conclusion}.

\section{Materials \& Methods}

This study explores the consistency of touch-based expression of emotions and touch gestures toward the robot by integrating tactile and auditory signals.
\label{sec2:Materials}

\subsection{Hardware materials}

\begin{figure*}[t]
\centering
\begin{subfigure}[t]{0.4\textwidth}
    \centering
    \includegraphics[height=6cm]{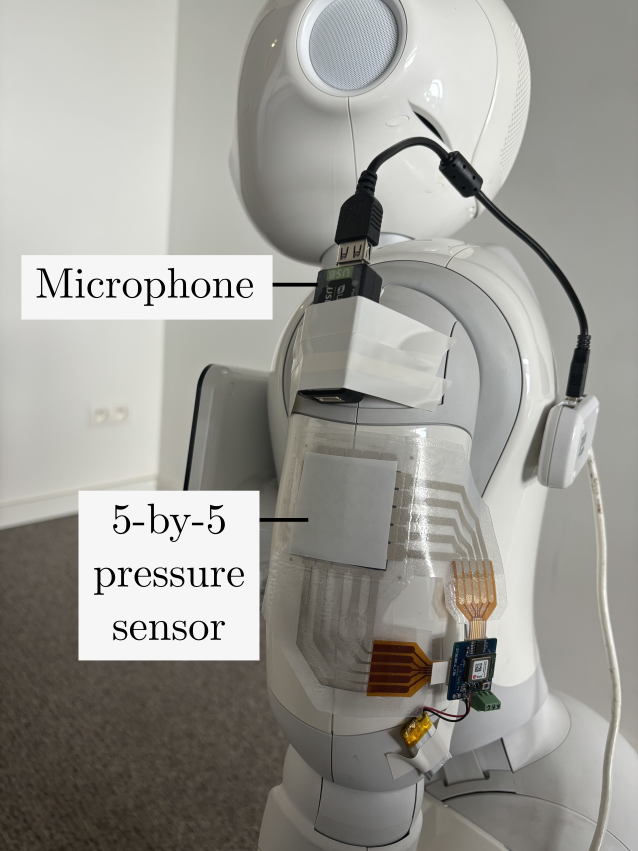}
\caption{Pepper robot mounted with sensors.}
    \label{fig:pepper_closeup}
\end{subfigure}
\begin{subfigure}[t]{0.55\textwidth}
    \centering
    \includegraphics[height=6cm]{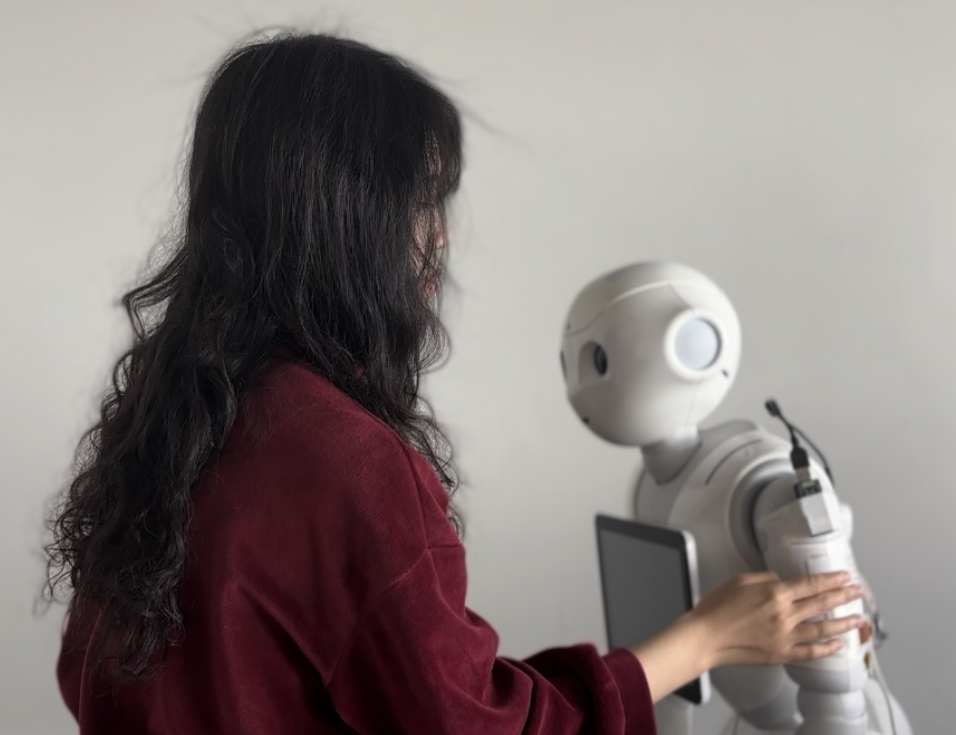}
    \caption{Participant interacting with Pepper.}
    \label{fig:pepperQQ}
\end{subfigure}
\caption{Experimental setup.}
\label{fig:experimentalsetup}
\end{figure*}

In this experiment, participants were tasked with conveying emotions to a Pepper robot, see Fig.~\ref{fig:experimentalsetup}.
The participants were presented with definitions of the emotions to ensure they agreed with and understood the given definitions.
The data collection involved the use of a custom 5-by-5 tactile sensor and a microphone to capture gesture data. 
Fig.~\ref{fig:pepper_closeup} shows how the sensors are mounted to the Pepper robot.
The following sections will detail the equipment used, the setup process, and the specific configurations implemented for data acquisition.


\subsubsection{Tactile sensor}

\begin{figure*}[t]
\centering
\begin{subfigure}[t]{0.4\textwidth}
    \centering
    \includegraphics[height=4cm]{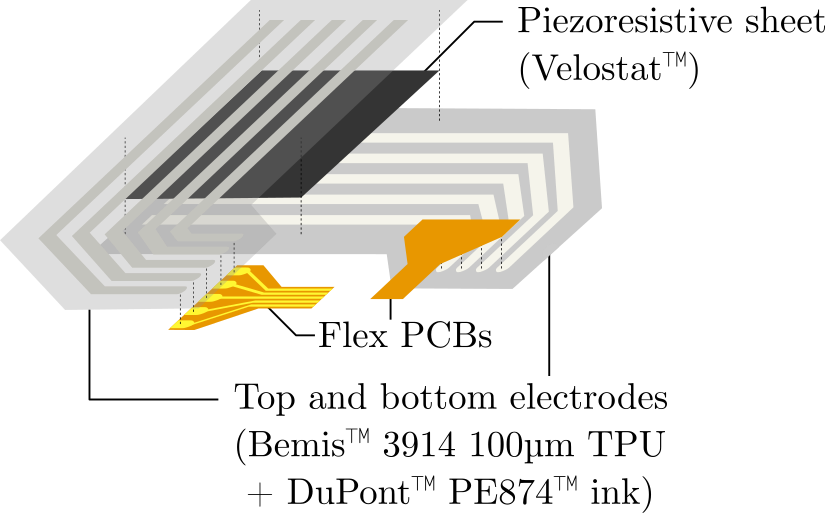}
\caption{Layered structure. Two extra layers of TPU cover the otherwise exposed traces, omitted for clarity.}
    \label{fig:sensor_exploded}
\end{subfigure}\hspace{5mm}
\begin{subfigure}[t]{0.4\textwidth}
    \centering
    \includegraphics[height=3.7cm]{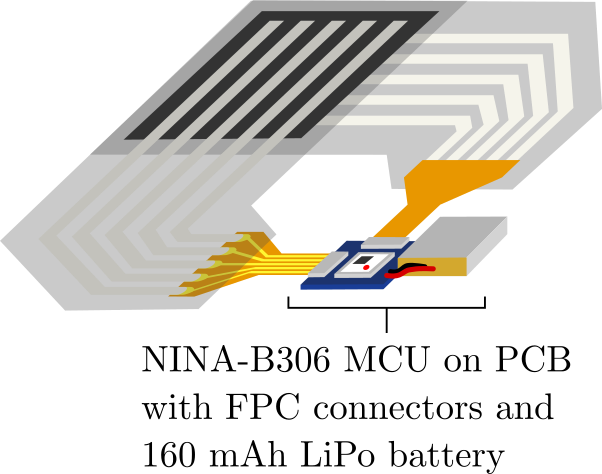}
    \caption{Complete sensor with readout electronics.}
    \label{fig:sensor_collapsed}
\end{subfigure}

\vspace{2mm}
\begin{subfigure}[t]{0.5\textwidth}
    \centering
    \includegraphics[height=4.3cm]{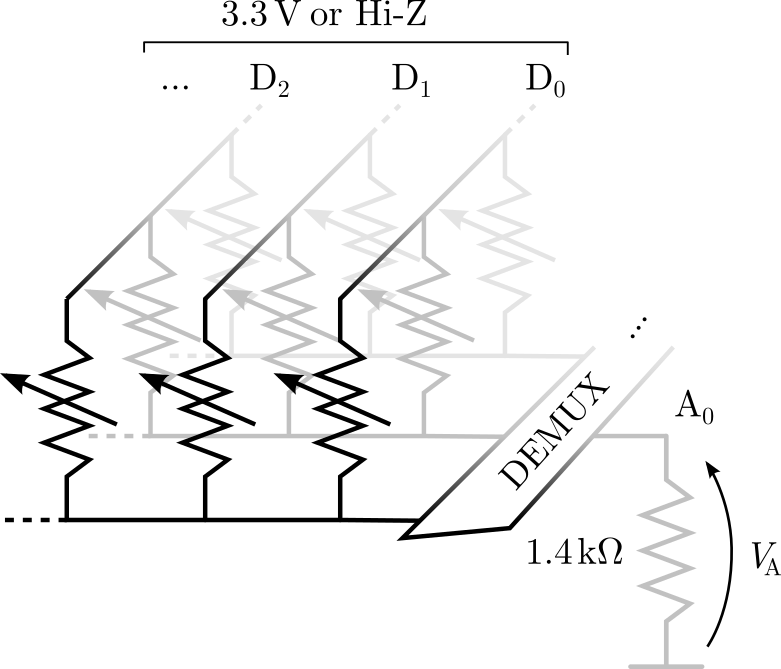}
    \caption{Equivalent circuit. D$_{0..5}$ and A$_0$ are digital and analog pins of the readout MCU.} 
    \label{fig:sensor_network}
\end{subfigure}
\caption{Structure of the tactile sensor.}
\label{fig:sensor}
\end{figure*}

The tactile sensor is a 5-by-5 piezoresistive pressure grid, similar to the Smart Textile developed in~\cite{proesmans2022smarttextile}.
As shown in Fig~\ref{fig:sensor_exploded}, the sensor consists of a top and bottom electrode pattern with a piezoresistive Velostat\texttrademark{} sheet in between.
The Velostat\texttrademark{} sheet was roughened by pressing it between two pieces of 80~grit sanding paper: this ensures little contact between the Velostat\texttrademark{} and the electrodes when no pressure is applied, and thus a small nominal sensor reading.
The electrodes are made by screenprinting DuPont\texttrademark{}~PE874\texttrademark{} conductive ink onto a \SI{100}{\micro\metre} Bemis\texttrademark{}~3914 thermoplastic polyurethane (TPU) sheet.
For each electrode, the printed side is facing the Velostat\texttrademark{} sheet.
To interface the printed TPU with a printed circuit board (PCB) for readout, flexible PCBs are glued to the printed TPU sheet.
This is achieved by adding extra conductive ink on the traces and Gorilla\textsuperscript{\tiny{\textregistered}} Super Glue in between.
Finally, extra TPU is used to cover the remaining exposed traces, to prevent oxidisation. 
This was omitted from Fig.~\ref{fig:sensor} for clarity of the graphic.
The entire structure is fused together using a Hotronix\textsuperscript{\tiny{\textregistered}} Air Fusion IQ\textsuperscript{\tiny{\textregistered}} heat press at \SI{120}{\degreeCelsius} and 1.4\,bar for \SI{10}{s}.

The readout PCB, shown in Fig.~\ref{fig:sensor_collapsed}, uses a NINA-B306 microcontroller unit (MCU), powered by an external \SI{160}{mAh} lithium-ion polymer (LiPo) battery.
Flexible printed cable (FPC) connectors are used to interface with the flex PCBs.
The MCU communicates the sensor data wirelessly over Bluetooth Low Energy (BLE) at an average rate of 45\,Hz.
Custom Arduino board files were developed for the PCB so that it can be programmed using the Arduino IDE. 
We also provide a guide for creating such a custom Arduino board in this GitHub repository\footnote{\url{https://github.com/RemkoPr/airo-nrf52840-boards}}.

The transduction mechanism of the sensor is as follows.
When pressure is applied to the sensor, the resistance of the Velostat\texttrademark{} sheet decreases.
Hence, as shown in Fig.~\ref{fig:sensor_network}, the sensor can be modelled as a grid of variable resistors.
Each of the traces of the top electrode is connected to a digital pin D$_{0..5}$ of the MCU.
The traces of the bottom electrode run into a demultiplexer (DEMUX), of which the output is connected to an analog pin A$_0$ of the MCU.
This pin has a 1.4\,k$\Omega$ pulldown resistor.
Such a low pulldown value biases the sensor as a sensitive touch sensor, rather than a pressure sensor with a large dynamic range.
To read a single taxel, one of the digital pins is set high (3.3\,V), the others to a high impedance (Hi-Z) state. 
The DEMUX then routes one of the bottom traces to the A$_0$ pin.
The voltage $V_A$ at this pin depends on the pulldown resistor and the resistance $R_x$ of the Velostat\texttrademark{} squeezed between the top trace set to a voltage level HIGH, and the bottom trace selected by the DEMUX:

\begin{equation}
    V_A = \frac{\SI{1.4}{k\ohm}}{\SI{1.4}{k\ohm} + R_x}\SI{3.3}{V}.
\end{equation}

\subsubsection{Microphone}

We used a Waveshare USB to audio microphone. 
The recordings were conducted in mono at a sample rate of 44.1\,kHz. 
The audio stream was processed in chunks of 1024 frames per buffer. 
Each recorded sample, lasting 10 seconds, was saved in WAV format.


\subsection{Experimental design}
To explore emotion decoding via touch-based expression and reveal how consistent emotions and gestures are expressed across different individuals, we set up a data collection experiment, which can be seen in Fig.~\ref{fig:experimentalsetup}. We recruited all the participants from a similar cultural background. Specifically, the group comprises 28 Chinese participants, 10 identified as female, 18 as male, and their average age is $27.8\pm 2.3$ years. The data collection and study adhered to the ethics procedures of Ghent University, and participants gave informed consent. 

\subsubsection{Definitions of emotions}

\begin{figure}
    \centering
        \includegraphics [width=\columnwidth]{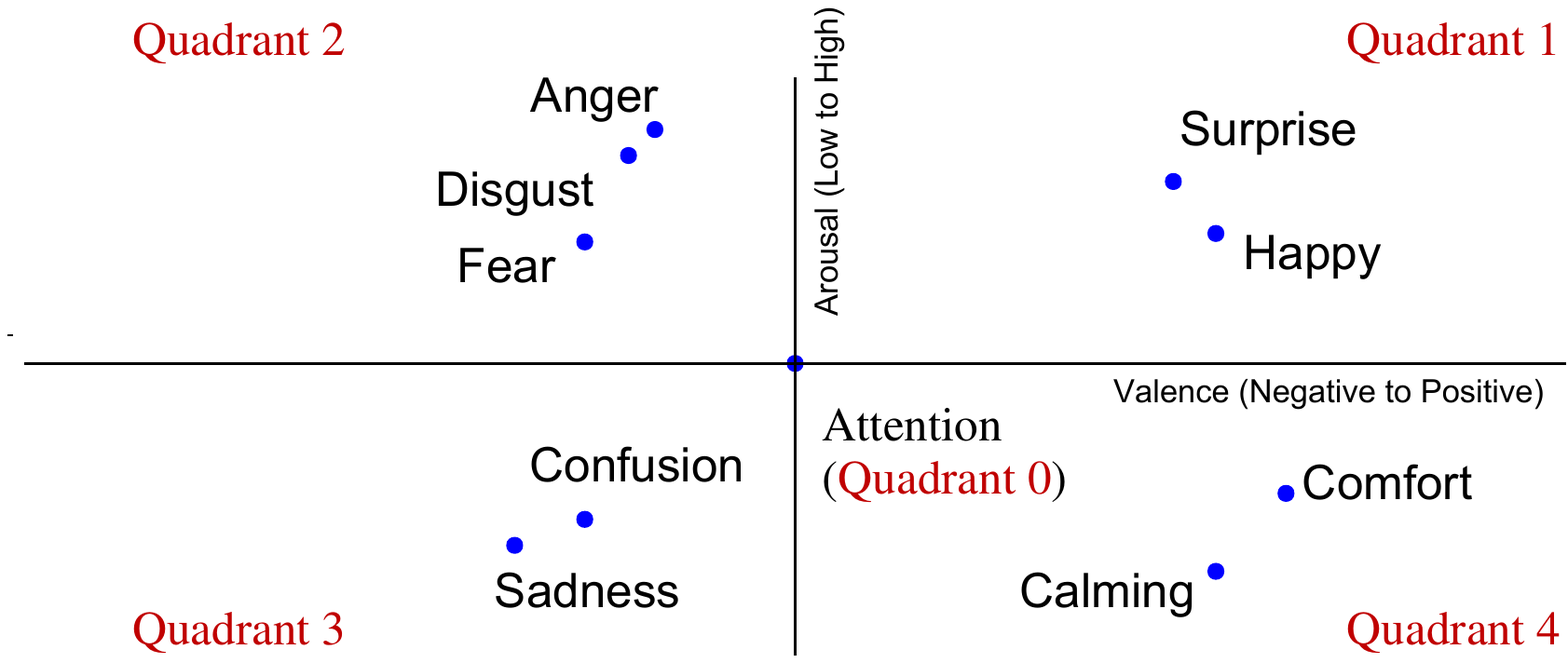}
    \caption{Russell's Circumplex Model of emotions~\cite{russell1989cross}.}
    \label{russel_model}
\end{figure}

Affective states can be represented along dimensions of arousal and valence, as described by Russell's model~\cite{russell1989cross}. In our study, we selected nine distinct emotions from different arousal and valence zones, plus one neutral emotion, 
as illustrated in Fig.~\ref{russel_model}, \enquote{Anger}, \enquote{Fear}, and \enquote{Disgust} fall within the high arousal and negative valence quadrant (Quadrant 2). \enquote{Happiness} and \enquote{Surprise} are in the high arousal and positive valence quadrant (Quadrant 1). \enquote{Sadness} and \enquote{Confusion} belong to the low arousal and negative valence quadrant (Quadrant 3), while \enquote{Comfort} and \enquote{calming} are from the low arousal and positive valence quadrant (Quadrant 4). \enquote{Attention} is categorized as a neutral emotion (Quadrant 0). Before data collection, participants were provided with definitions of these ten emotions to ensure they understood and agreed with the interpretations or could offer their own understandings. Details of each emotion definition refer to the Oxford dictionary and can be found at this GitHub repository\footnote{\url{https://github.com/qiaoqiao2323/Convey-emotion}}.

\subsubsection{Data collection procedure}
The experiment followed a within-subject design using two sessions: A) \textit{Conveying emotions to the robot}, and B) \textit{Conveying tactile gestures to the robot}.

\begin{enumerate}[label=(\Alph*)]
    \item \textit{Conveying emotions to the robot:} In this session, participants were asked to convey each of the 10 emotions to the robot, one at a time, using spontaneous gestures, with each gesture lasting 10 seconds. 
    
    \item \textit{Conveying tactile gestures to the robot:} In this session, participants were instructed to convey each of the six gestures to the robot, one at a time, with each gesture lasting 10 seconds. 
    
\end{enumerate}

Participants were given time between each emotion and gesture to prepare and think about their next performance. Subsequent data collection only commenced once the participants indicated they were ready. Participants completed three rounds, with each subsequent round commencing only after all emotions or gestures had been performed, ensuring a greater variety in the data collected.
To prevent bias or influence on the gestures used by participants to convey emotions, session A was consistently conducted prior to session B. 

\subsubsection{Questionnaire} \label{ss:questionnaire}

After collecting gesture data, we gathered subjective feedback through a questionnaire aimed at understanding the challenges of affective touch expression and the correlation between affective touch and tactile gestures:

\begin{itemize}
    \item \textit{Challenging emotions to express:} We asked participants which emotions or intentions they find more challenging to express through a single multiple-choice question: ``Which emotions or intentions do you find most difficult to express?''
    
    \item \textit{Similar emotions based on touch expression:}  We gathered pairs of similar emotions by asking participants: ``Are there any emotions listed below that you think are similar, based solely on how you express them? Please list your answers in pairs, such as Emotion A -- Emotion B.''

    \item \textit{Correlations between emotions and gestures:} We examined the correlations between emotions and gestures by asking: ``Do you believe that certain gestures are associated with specific emotions? If so, which gestures correspond to which emotions?''
    
    \item \textit{General feedback:} Finally, we pose an open question: \enquote{Do you have suggestions or feedback for this experiment?}

\end{itemize}

\subsection{Data processing}
\label{sec:data_processing}

Data processing and analysis were done using Python and R V4.3.2. Audio and tactile features were extracted using Python and then analyzed using R. A Permutational Multivariate Analysis of Variance (PERMANOVA) analysis was used to compare data between conditions. A Shapiro-Wilk analysis (with $\alpha = 0.05$) was used to confirm the normality of distributions. 

\subsubsection{Dataset}

The dataset consists of tactile sensor and microphone recordings collected to study the conveyance of emotions through touch and audio. 
For each of the 10 emotions, there are a total of 84 samples (28 participants for 3 rounds). Each interaction lasted 10 seconds. Tactile sensor data were sampled at a rate of 45\, Hz, providing continuous feedback during the interactions. Simultaneously, 44.1\,kHz audio data were recorded in 10-second WAV files.

\subsubsection{Audio Features}
Inspired by~\cite{shoiynbek2019robust}, the audio features extracted in this study include Mel-Frequency Cepstral Coefficients (MFCCs), Spectral Centroid, Spectral Bandwidth, Zero Crossing Rate, and Root Mean Square Energy (RMSE). MFCCs (mfcc\_1 to mfcc\_13) are critical in capturing the timbral qualities of touch sounds, making them effective for distinguishing different emotional tones conveyed through touch. 
The Spectral Centroid represents the \enquote{center of mass} of the spectrum, correlating with the perceived brightness of a sound; higher values often suggest more energetic emotions like happiness or excitement. 
Spectral Bandwidth measures the range of frequencies present in a sound, with broader bandwidths typically indicating more complex or intense touch sounds. 
The Zero Crossing Rate, which counts the number of times the audio signal changes sign, can indicate the noisiness or smoothness of a touch sound and is useful for detecting emotions like agitation or calmness. 
RMSE provides an overall measure of the energy present in the audio signal, with higher values potentially signifying stronger emotions such as anger or enthusiasm.

\subsubsection{Tactile Features}
Inspired by~\cite{hertenstein2006touch}, tactile features extracted from each sample include Mean Pressure, Max Pressure, Pressure Variance, Pressure Gradient, Median Force, Interquartile Range Force (IQR Force), Contact Area, Rate of Pressure Change, Pressure Standard Deviation (Pressure Std), Number of Touches (Num\_touches), Max Touch Duration, Min Touch Duration, and Mean Duration. Mean Pressure provides an average measure of the applied pressure, while Max Pressure identifies the peak force applied during touch interactions, both of which can reflect the intensity of emotions conveyed. Pressure Variance and Pressure Gradient offer insights into the consistency and change rate of the applied pressure, respectively, with high variability or gradient often associated with dynamic emotions like anxiety or excitement.

Median Force, representing the middle value of force applied, IQR Force measures the dispersion of the middle 50\% of force values. Contact Area is crucial in identifying the nature of the touch, with larger areas potentially indicating comforting emotions, while smaller areas might suggest precision or nervousness. The Rate of Pressure Change captures how quickly the pressure varies, with high rates linked to emotions like excitement. Pressure Std reflects the fluctuation in pressure, and Num\_touches quantifies the frequency of touch events within 10 seconds. Max Touch Duration, Min Touch Duration, and Mean Touch Duration describe the temporal aspects of touch, indicating how long touches are sustained, which can be correlated with different emotional expressions.

\section{Data analysis}
\label{data_analysis}

\subsection{Subjective Feedback Analysis on Touch-based Emotional Expression}

In our experiment, participants provided subjective feedback on emotional expression through touch. These areas included challenging emotions for expression, similarities in emotional expression based on touch, and correlations between emotions and gestures.

\subsubsection{Challenging Emotions for Expression}
Participants found certain emotions more challenging to express through touch. The results, as shown in Fig.~\ref{fig:challenging_emotion}, indicate that \enquote{Surprise} and \enquote{Confusion} were the most difficult emotions to convey. Specifically, 64.3\% of participants identified \enquote{Confusion} as challenging, while 46.4\% reported difficulty expressing \enquote{Surprise}. These emotions stood out among the ten evaluated in the study. On the contrary, no participants regarded \enquote{Attention} as challenging to convey.

\subsubsection{Similar Emotions Based on Touch Expression}
When asked to identify emotions that felt similar based on touch expression alone, participants highlighted several pairs. Fig.~\ref{fig:similar_emotions} shows that 14 participants chose ``Calming'' and ``Comfort'' as similar emotions. These emotions were consistently associated with low-frequency touch movements and light force. Additionally, 9 participants found ``Disgust'' and ``Anger'' to be similar, likely due to both emotions being characterized by high arousal and negative valence. We found that all the similar affective touch pairs given by the participants always belong to the same arousal or sample valence zone; the lack of social context in the experiment may have contributed to the difficulty in differentiating these emotions through touch alone.

\subsubsection{Correlations Between Emotions and Gestures}
\begin{figure*}[t]
\centering
\begin{subfigure}[t]{0.52\textwidth}
    \centering
    \includegraphics[width=\columnwidth]{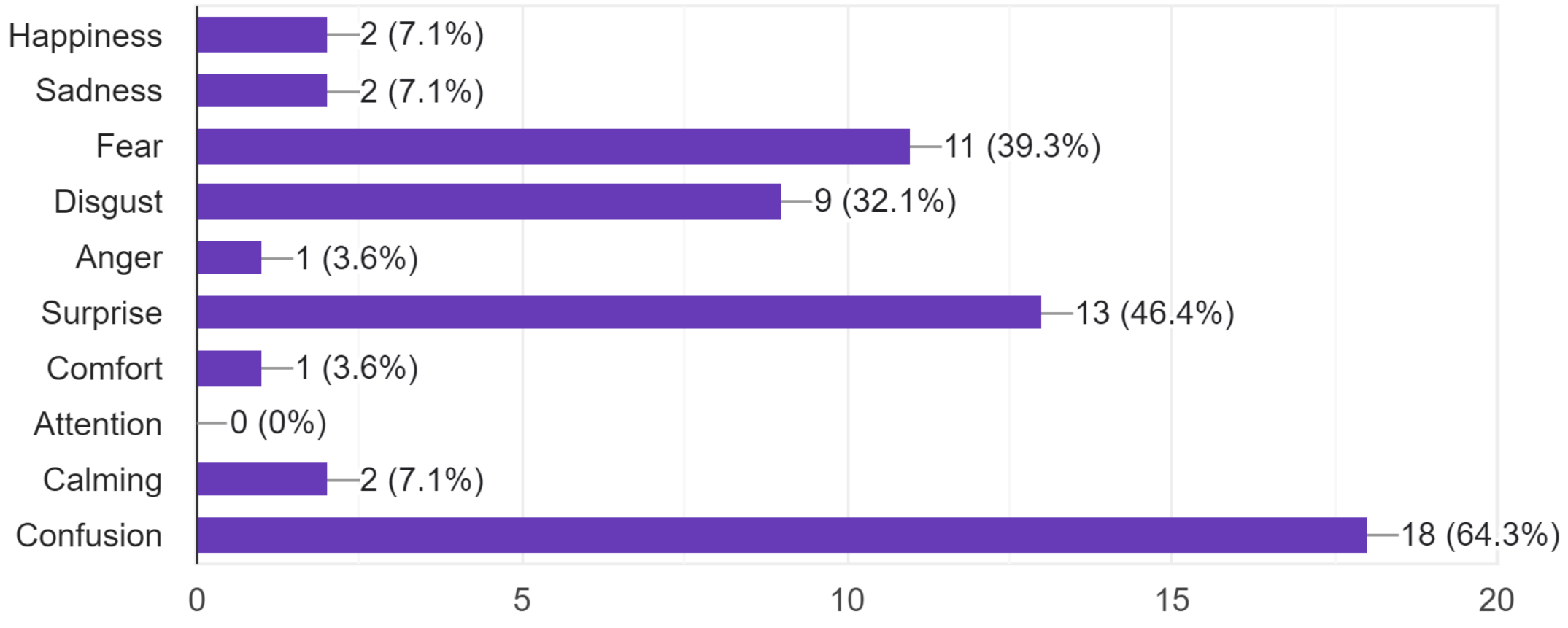}
\caption{Challenging emotions to express through touch gestures. This bar chart depicts the number and percentage of participants who found specific emotions challenging to express through touch gestures. 
}
    \label{fig:challenging_emotion}
\end{subfigure}\hfill
\begin{subfigure}[t]{0.43\textwidth}
    \centering
    \includegraphics[width=\columnwidth]{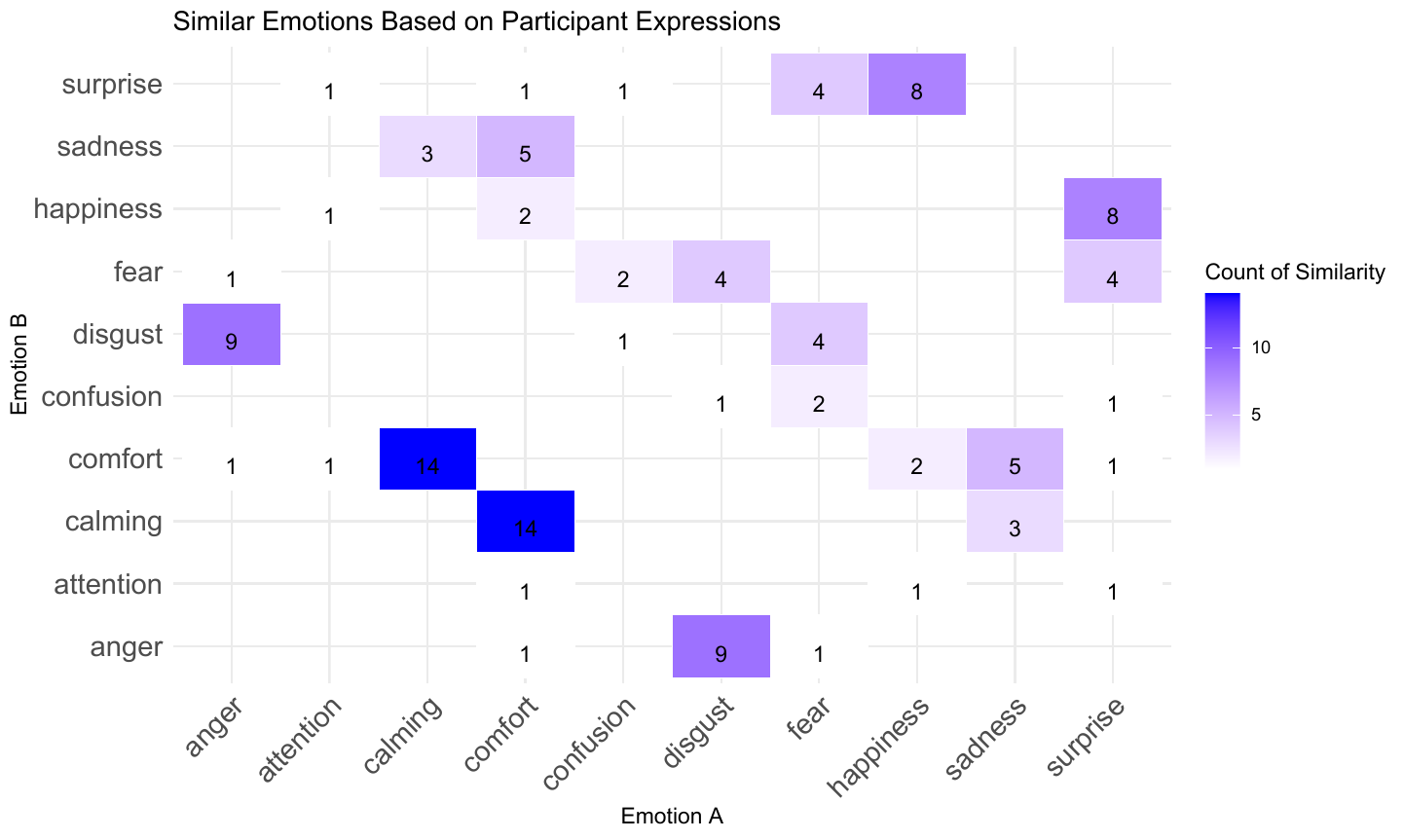}
    \caption{Confusion matrix showing the similarities between different emotions based on participants' expressions. Each cell in the matrix represents the count of instances where participants expressed emotion pairs to be similar.}
    \label{fig:similar_emotions}
\end{subfigure}
\begin{subfigure}[t]{\textwidth}
    \centering
        \includegraphics [width=\columnwidth]{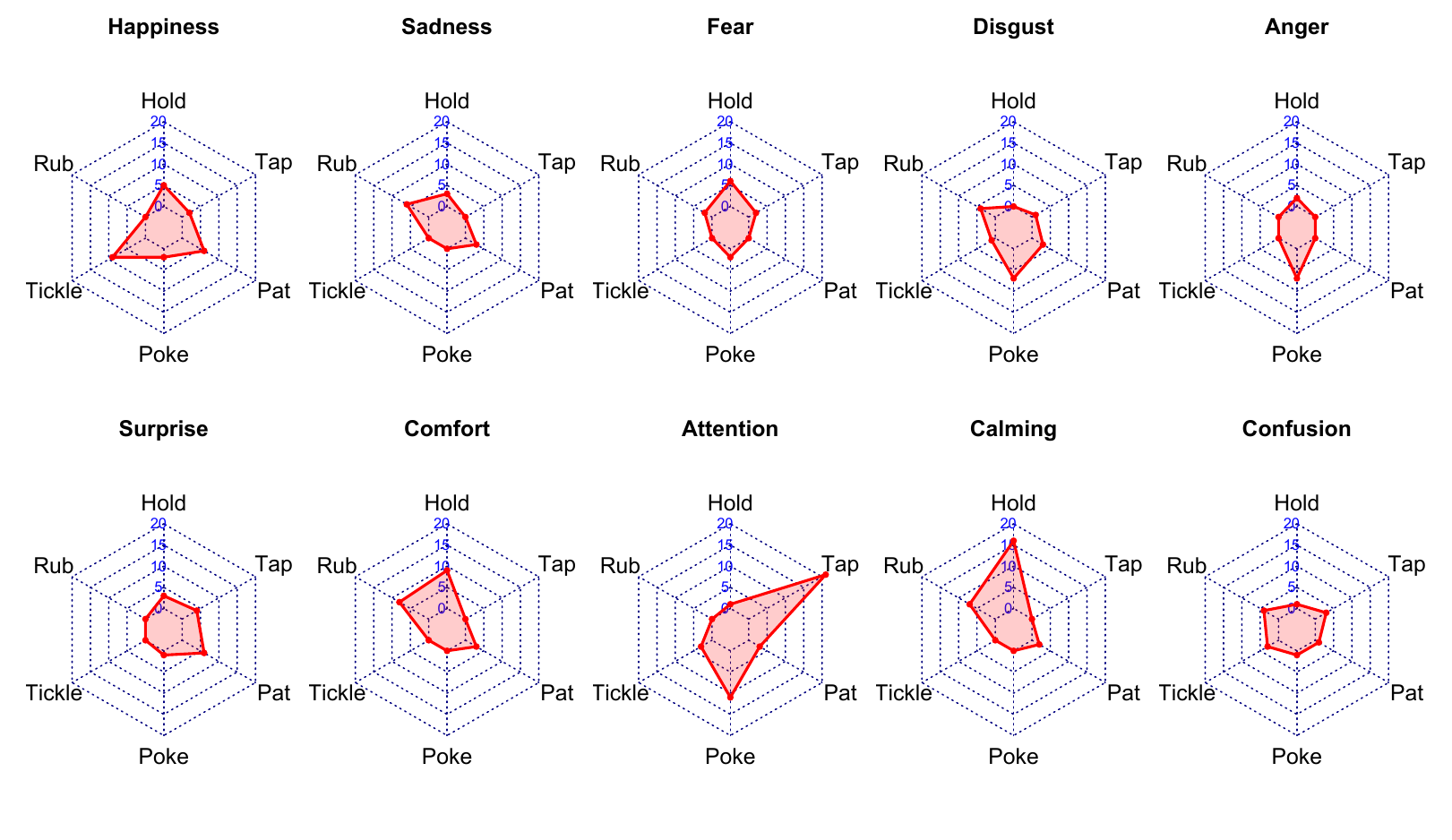}
    \caption{Correlations between touch gestures and emotions based on participants' subjective feedback.
    The radar charts show the number of times each gesture was indicated to be appropriate for a given emotion by the participants. Some emotions like ``Attention'' have a higher association with  ``Tap'' gestures, while some emotions like ``Disgust'' and ``Confusion'' do not show an obvious correlation with gestures.}
    \label{fig:gesture_emotion}
\end{subfigure}
\caption{Questionnaire results.}
\label{fig:questionnaire_results}
\end{figure*}

Participants’ feedback revealed interesting correlations between emotions and specific gestures. As illustrated in Fig.~\ref{fig:gesture_emotion}
, each emotion could be linked to multiple gestures, suggesting a flexible strategy for expressing emotions. For instance, 16 participants chose holding as a gesture for \enquote{Calming}, while 7 participants preferred \enquote{Rubbing}. Surprisingly, 21 participants associated tapping with ``Attention'', and 11 participants chose ``Poking''. ``Confusion'' was linked to various gestures, but none were strongly correlated. On the other hand, ``Attention'', ``Comfort'', and ``Calming'' showed stronger associations with specific gestures.

\subsubsection{Suggestions and feedback for the experiment}
Participants provided valuable suggestions for improving the experiment. Some recommend increasing the touch area to enhance emotional expression. Feedback also indicated that considering social roles and context, such as gender, relationship, and appearance, could influence how emotions are conveyed. Participants suggested varying the touch positions and incorporating external stimuli for a more comprehensive analysis. Specific gestures were also highlighted: two participants mentioned using ``Stroking'' for ``Comfort'' and ``Sadness'', six indicated hitting to express ``Anger'', and two suggested pushing. Additionally, several participants expressed concerns about the force used on the robot when expressing ``Anger'' and ``Disgust'', fearing potential damage, and requested guidelines on the safe amount of force to apply during the experiment.

The feedback highlights the complexity of expressing emotions through touch. The varying difficulty in conveying different emotions and the identification of similar arousal and valence zone emotions based on touch alone is challenging. The correlations between emotions and gestures suggest that participants use a combination of gesture features, such as amplitude, frequency, and contact area, to convey different emotions effectively.


\subsection{Consistency analysis}

In order to transform feature values to a similar scale, ensuring all features contribute equally to the model, we standardized the tactile and audio features. After standardization, the numeric features of the dataset have a mean of 0 and a standard deviation of 1. We then performed Principal Component Analysis (PCA) to reduce dimensionality and identify the key components capturing the most variance. The resulting principal component scores were integrated with the original emotion and participant identifiers. To evaluate the consistency of emotion conveyance, we calculated Intraclass Correlation Coefficients (ICC) for each emotion using the PCA scores and a two-way consistency model.

\begin{figure*}
\centering
\begin{subfigure}[t]{0.52\textwidth}
    \centering
    \includegraphics[width=\columnwidth]{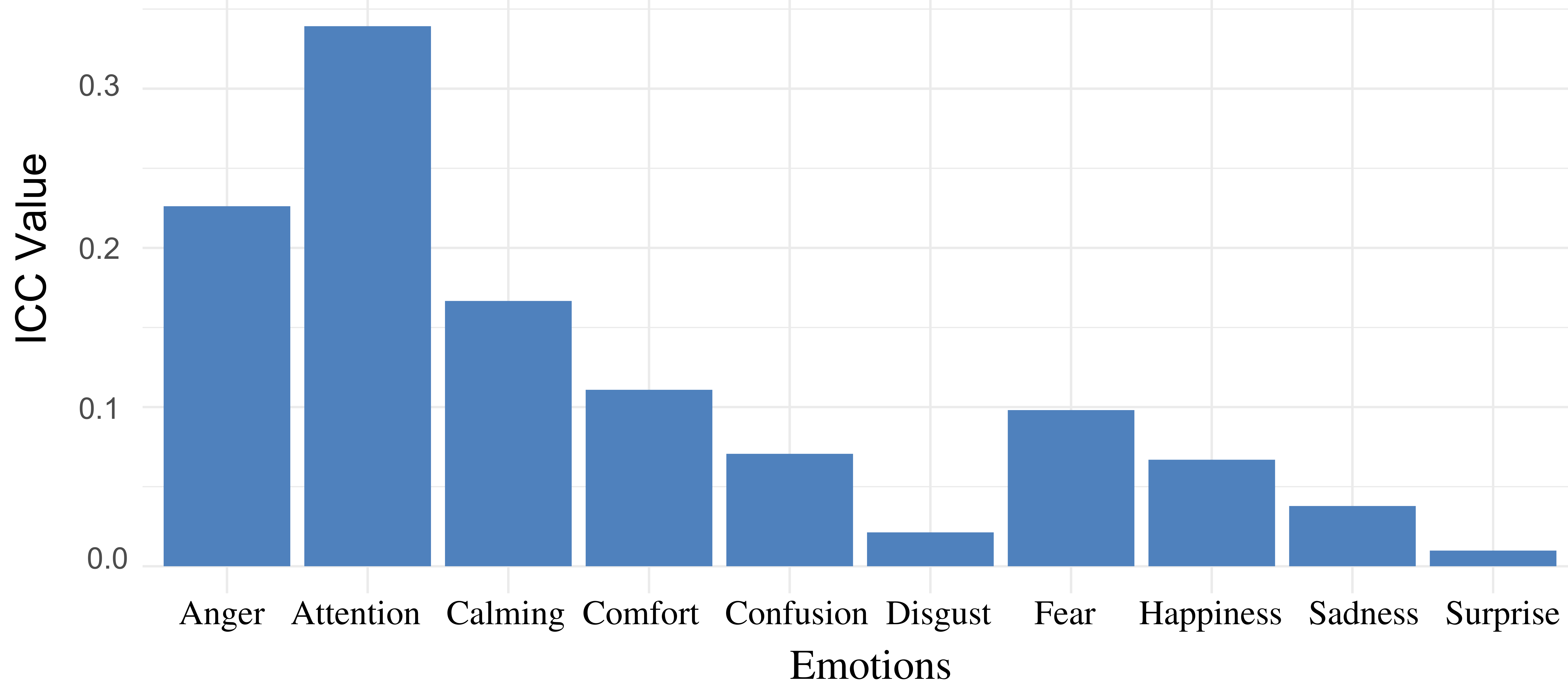}
\caption{ICC values among emotions, the bars show the ICC values of each emotion}
    \label{fig:icc_emotions}
\end{subfigure}
\begin{subfigure}[t]{0.43\textwidth}
    \centering
    \includegraphics[width=\columnwidth]{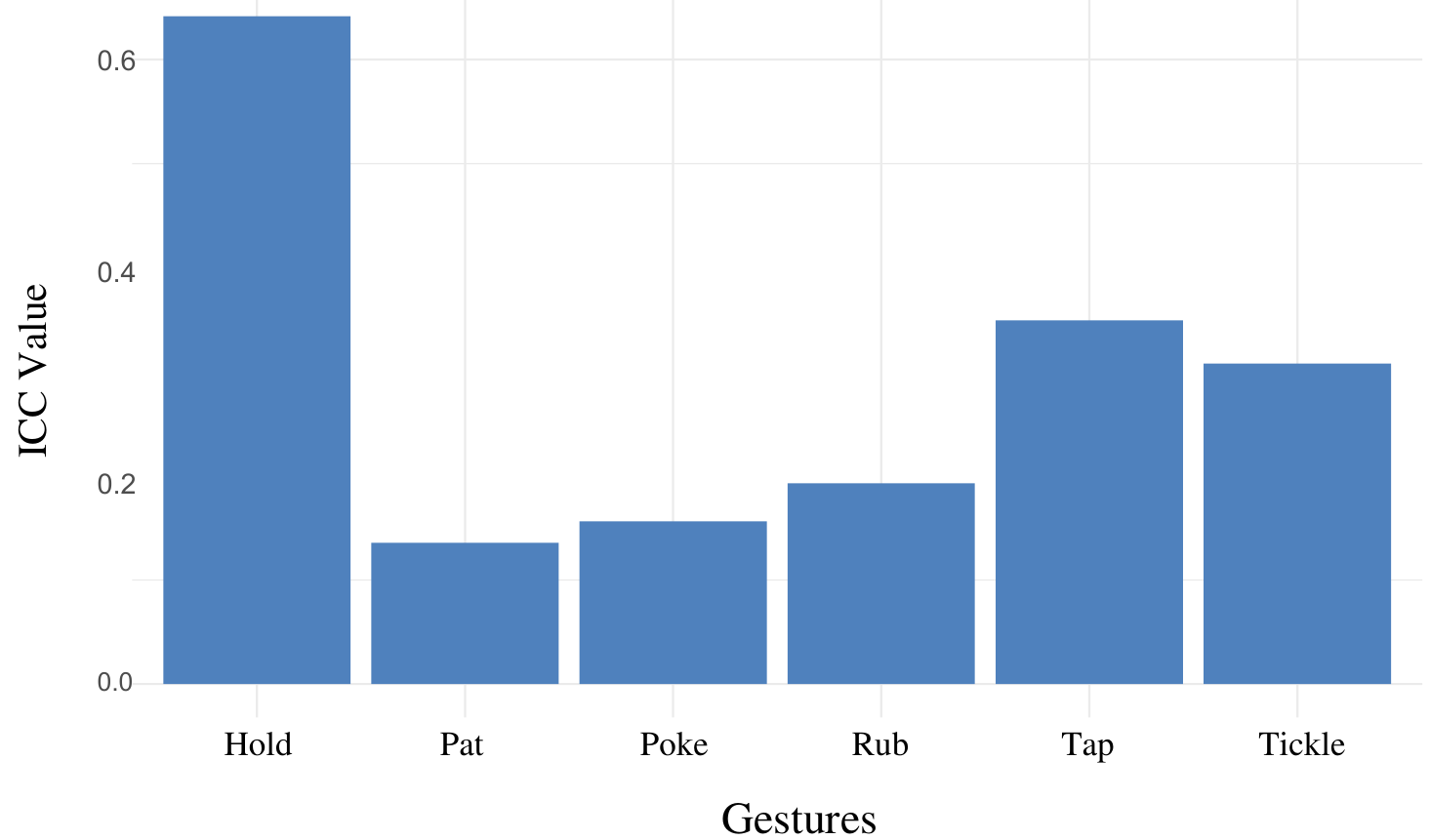}
    \caption{ICC values among gestures.}
    \label{fig:icc_gestures}
\end{subfigure}
\caption{Consistency analysis.}
\label{fig:icc}
\end{figure*}

As shown in Fig.~\ref{fig:icc_emotions} and Table~\ref {tab:ICC}, all emotions demonstrated statistically significant consistency across participants, although some emotions exhibited comparatively low ICC values. The highest ICC was observed for \enquote{Attention}, indicating the most consistent expression, while \enquote{Surprise} had the lowest ICC, suggesting the greatest variability in expression. 




Fig. \ref{fig:icc_gestures} and Table \ref{tab:ICC_gesture} demonstrate that all gestures exhibit significant consistency across participants. The \enquote{Hold} gesture achieved the highest ICC at 0.64, whereas the \enquote{Pat} gesture had the lowest ICC at 0.14. The \enquote{Tap} and \enquote{Tickle} gestures showed similar ICC values, at 0.35 and 0.31, respectively.

\begin{table*}\footnotesize 
\setlength{\abovecaptionskip}{0.0cm}   
	\setlength{\belowcaptionskip}{-0cm}  
	\renewcommand\tabcolsep{2.0pt} 
	\centering
	\caption{Intraclass Correlation Coefficients among participants (** at p < 0.001).}
	\begin{tabular}
	{
	p{1.5cm}<{\centering} 
 p{1.9cm}<{\centering} 
	 p{1.9cm}<{\centering} 
  p{1.9cm}<{\centering}
	p{1.9cm}<{\centering}
 	p{2.2cm}<{\centering}
	} 
\hline
    
     {Emotions} & 
     {Anger}  & 
     {Attention} & 
     {Calming} & 
     {Comfort} & 
     {Confusion}  \\

     \cline{1-6}

   {ICC} &
  {0.23 **} &
 {0.34 **} &
 {0.17 **} &
  {0.11 **} &
 {0.07 **} 
\vspace{3mm}\\

     \hline

 {Emotions} & 
     {Disgust} & 
     {Fear} & 
     {Happiness} & 
     {Sadness} & 
     {Surprise} \\

     \cline{1-6}

   {ICC} &
  {0.02 **} &
 {0.10 **} &
 {0.07 **} &
  {0.04 **} &
 {0.01, $p = 0.004$} \\

	\end{tabular}
	\label{tab:ICC}
\end{table*}


\begin{table*}[t]
\footnotesize 
\setlength{\abovecaptionskip}{0.0cm}   
	\setlength{\belowcaptionskip}{-0cm}  
	\renewcommand\tabcolsep{2.0pt} 
	\centering
	\caption{ICC of tactile gestures among participants, the bars show the ICC values of each gesture.}
	\begin{tabular}
	{
	p{1.5cm}<{\centering} 
 p{2.2cm}<{\centering} 
	 p{2.2cm}<{\centering} 
  p{2.2cm}<{\centering}
	p{2.2cm}<{\centering}
 	p{2.2cm}<{\centering}
 	p{2.2cm}<{\centering}
	} 
\hline

     {Emotions} & 
     {Hold} & 
     {Pat}  & 
     {Poke} & 
     {Rub} & 
     {Tap} & 
     {Tickle} \\

\hline
   {ICC} &
  {0.64, $p < 0.001$} &
  {0.14, $p < 0.001$} &
 {0.16, $p < 0.001$} &
  {0.19, $p < 0.001$} &
  {0.35, $p < 0.001$} &
 {0.31, $p < 0.001$} \\
 \hline

	\end{tabular}
	\label{tab:ICC_gesture}
\end{table*}


\subsection{Variability analysis}

\subsubsection{Variability analysis emotional touch expression}
\label{subsubsection:affective touch}

The features do not follow a normal distribution. 
Therefore, we used permutational multivariate analysis of variance (PERMANOVA), which reveals a significant effect of emotion on the scaled combined features ($F = 14.06, p < 0.001$). 
Pairwise comparisons with Holm correction show that most emotion pairs differ significantly and have consistent significance after p-value adjustment ($p<0.05$). However, there is no difference between \enquote{Calming} and \enquote{Sadness} ($F = 3.78, p = 0.09$), \enquote{Comfort} and \enquote{Sadness} ($F = 4.89, p = 0.09$) based on touch and audio features, and there is no difference between \enquote{Confusion} and \enquote{Comfort} ($F = 4.18, p = 0.14$), \enquote{Confusion} and \enquote{Sadness} ($F = 1.72, p = 1.00$), \enquote{Disgust} and \enquote{Sadness} ($F = 3.21, p = 0.36$), \enquote{Disgust} and \enquote{Surprise} ($F = 1.84, p = 1.00$), \enquote{Happiness} and \enquote{Surprise} ($F = 2.84, p = 1.00$). We hereby analyse the four valence and arousal quadrants plus one neutral emotion \enquote{Attention}, and the pairwise comparison with post-hoc analysis shows that all the pairwise comparison has significant differences between each other ($p<0.05$) except for Quadrant 2 and Quadrant 3 ($p = 0.06$), which is close to the significant level.  

Intuitively, it is expected that \enquote{Surprise} is very similar to \enquote{Happiness}. One explanation could be that both of the emotions belong to the high-arousal zone. Similarly, there is no significant difference between \enquote{Confusion} and \enquote{Sadness}, or between \enquote{Disgust} and \enquote{Sadness}, as they belong to the low-arousal zone.

\subsubsection{Variability analysis for tactile gesture}

Following the same procedure in affective touch variability analysis, refer to Subsection~\ref{subsubsection:affective touch}, the features do not follow a normal distribution. Therefore, we used PERMANOVA, which reveals a significant effect of gesture on the scaled combined features ($F = 99.33, p < 0.001$). Pairwise comparisons with Holm correction show that all the gesture pairs differ significantly and have consistent significance after p-value adjustment ($p<0.05$).



\section{Decoding emotions using different modalities} 
\label{s:decoding_emotions}

\subsection{Data split} 
We split the training and test datasets based on participant IDs to avoid data leakage. 
Specifically, we randomly sampled 22 participants for the training set and 6 participants for the test set. All the extracted features, code, and visualization demos are available on GitHub\footnote{\url{https://github.com/qiaoqiao2323/Convey-emotion}}.

\subsection{Model types} 
We used various machine learning techniques to decode the emotions conveyed by touch and sound.
\subsubsection{Classical machine learning models}
We selected Random Forest (RF)~\cite{breiman2001random}, Support Vector Machine (SVM)~\cite{hearst1998support}, and Decision Tree (DT)\cite{freund1999alternating} for their robustness and proficiency in handling high-dimensional data. Specifically, RF excels at handling high-dimensional data and has built-in feature importance for classification~\cite{breiman2001random, liaw2002classification}. Decision trees are generally well-suited for small datasets and have been used for multimodal classification in~\cite{chen2004decision, brown1993comparison}. SVM has been used in touch gesture classification and audio emotion classification~\cite{jung2017automatic, chen2012speech}. The models were evaluated using 10-fold cross-validation to ensure robustness and accuracy. The results on the test dataset of these comparisons are presented in Table~\ref{tab:modelcomparisongestures}. For all traditional models, we performed a grid search to optimize key hyperparameters, such as the number of estimators and maximum depth for RF, kernel type and regularization parameter for SVM, and depth for DT, using 10-fold cross-validation to ensure generalizability and prevent overfitting. All model architectures, training code, and hyperparameters are publicly available in our GitHub repository\footnote{\url{https://github.com/qiaoqiao2323/Convey-emotion}}.

\subsubsection{Deep learning models}
In addition, deep learning is frequently used for emotion and touch gesture decoding. We explored five different models: a multi-temporal resolution convolutional neural network~\cite{hou2024sound} (MTRCNN), a pre\-trained audio neural network (PANN)~\cite{kong2020panns}, 
and several models which have already been used before for touch gesture classification, namely a Convolution Neural Network - Gated Recurrent Unit (CNN\_GRU), a CNN\ Transformer~\cite{darlan2023recognizing}, and a CNN\ LSTM.

Each model architecture is described below.
\begin{itemize}
    \item MTRCNN is a multi-task convolutional network, using a multi-branch architecture with different convolution kernel sizes (3×3, 5×5, 7×7). Each branch includes a convolutional block followed by dilated convolutions and pooling layers to extract multi-scale features. The outputs of these branches are concatenated and passed through a fully connected embedding layer, followed by a classification layer for the final predictions. 
\item CNN\_LSTM integrates convolutional layers with an LSTM to capture both spatial and temporal features. It uses three convolutional layers followed by a two-layer LSTM to model temporal dependencies, and the final output is obtained through a fully connected layer. 
\item CNN\_Transformer combines convolutional feature extraction with Transformer-based temporal modelling. The input is passed through three convolutional layers, and then processed by a Transformer encoder to capture long-range dependencies, and the final output is obtained through a fully connected layer. \item PANNs consists of six convolutional blocks, each progressively increasing feature dimensions. The network employs batch normalization, ReLU, dropout, and pooling within each block. After feature extraction, global frequency averaging and time aggregation are performed, followed by a fully connected layer for final class predictions. 
\item The CNN\_GRU model begins by applying three consecutive 3×3 convolutional layers. The number of output channels increases between layers, from 64 to 128, followed by 256. The GRU consists of two layers with a hidden size of 64 units, capturing the temporal dependencies in the input sequence. The GRU's output is flattened and passed through a fully connected layer to make the class predictions. 
\end{itemize}

\subsection{Model training} 
Each model is trained on sound features only, on tactile features only, and on both feature sets combined, to examine whether multimodal emotion decoding outperforms single-modality decoding. We compared model fusion and feature fusion techniques and found that feature fusion generally yielded better performance. This could be due to the increased complexity introduced by model fusion for our relatively small dataset. Consequently, we opted for feature fusion for multimodal decoding.
Results are averaged over 10 training runs.
\subsubsection{Classical machine learning models}
We use the Adam optimiser~\cite{kingma2014adam}, with a default learning rate of 0.0001. Each model trains for 100 epochs.

\subsubsection{Deep learning models}

NN models were trained for 100 epochs using a batch size of 32. Performance evaluations were conducted at the end of each epoch. The optimisation employed a standard configuration without AdamW, using a fixed learning rate of 0.001 throughout training.

\subsection{Decoding results \& discussion}

The results for gesture classification and emotion classification are given in Tables~\ref{tab:modelcomparisongestures} and \ref{tab:modelcomparisonemotions}, respectively.

\subsubsection{Emotion and touch gesture decoding} 

\begin{table*}[h]\footnotesize 
\setlength{\abovecaptionskip}{0.0cm}   
	\setlength{\belowcaptionskip}{-0cm}  
	\renewcommand\tabcolsep{2.0pt} 
	\centering
	\caption{Model comparison for gesture classification.}
	\begin{tabular}
	{
	p{3.5cm}<{\centering} 
 p{1.5cm}<{\centering} 
	 p{1.5cm}<{\centering} 
  p{2cm}<{\centering}
	p{2.5cm}<{\centering}
 	p{1cm}<{\centering}
 	p{1.5cm}<{\centering}
    	p{1.0cm}<{\centering}
 	p{1.0cm}<{\centering}
	} 
\hline
    
    {Models test accuracy} & 
    {MTRCNN}  & 
     {PANNs}  & 
     {CNN\_GRU} & 
     {CNN\_Transformer} & 
     {CNN\_LSTM} & 
      {RF} & 
     {SVM} & 
     {DT} \\

     \hline

   {Multimodal feature fusion} &
  {81.52\%} &
   {81.24\%} &
 {82.54\%} &
 {76.21\%} &
  {\textbf{90.74\%}} &
  {81.85\%} &
 {80.56\%} &
 {73.15\%} \\

  {Sound model} &
  {\textbf{78.89\%}} &
  {58.33\%} &
 {67.13\%} &
 {53.61\%} &
  {67.04\%} &
    {51.85\%} &
 {62.96\%} &
 {49.07\%} \\

  {Touch model} &
  {79.53\%} &
   {80.97\%} &
 {\textbf{81.39\%}} &
 {74.59\%} &
  {79.63\%} &
   {79.54\%} &
 {77.78\%} &
 {72.22\%} \\


     \hline
    
	\end{tabular}
\label{tab:modelcomparisongestures}
\end{table*}

\begin{table*}[h]\footnotesize 
\setlength{\abovecaptionskip}{0.0cm}   
	\setlength{\belowcaptionskip}{-0cm}  
	\renewcommand\tabcolsep{2.0pt} 
	\centering
	\caption{Model comparison for emotion classification.}
	\begin{tabular}
	{
	p{3.5cm}<{\centering} 
 p{1.5cm}<{\centering} 
	 p{1.5cm}<{\centering} 
  p{2cm}<{\centering}
	p{2.5cm}<{\centering}
 	p{1cm}<{\centering}
 	p{1.5cm}<{\centering}
    	p{1.0cm}<{\centering}
 	p{1.0cm}<{\centering}
	} 
\hline
    
    {Models test accuracy} & 
    {MTRCNN}  & 
     {PANNs}  & 
     {CNN\_GRU} & 
     {CNN\_Transformer} & 
     {CNN\_LSTM} & 
      {RF} & 
     {SVM} & 
     {DT} \\

     \hline

    {Multimodal feature fusion} &
{31.11\%} &
  {29.61\%} &
 {\textbf{40.00\%}} &
 {28.89\%} &
  {31.85\%} &
  {38.44\%} &
 {\textbf{40.00\%}} &
 {31.11\%} \\

  {Sound model} &
  {24.72\%} &
  {23.78\%} &
 {\textbf{28.33\%}} &
 {17.44\%} &
  {26.17\%} &
  {23.00\%} &
 {22.55\%} &
 {23.89\%} \\

  {Touch model} &
  {23.33\%} &
  {28.95\%} &
 {22.00\%} &
 {20.74\%} &
  {20.31\%} &
  {36.83\%} &
 {\textbf{36.11\%}} &
 {30.28\%} \\

     \hline
    
	\end{tabular}
\label{tab:modelcomparisonemotions}
\end{table*}



For the test data, both the SVM model with a linear kernel and the CNN\_GRU model showed the highest overall accuracy, each achieving 40.00\,\% in multimodal emotion decoding, which is significantly higher than the chance level of 10\,\% (p<0.05). Specifically, the SVM model scored 40\% accuracy when combining sound and touch modalities, 22.55\,\% with just sound, and 36.11\,\% with just touch. The CNN\_GRU model also reached 40\% accuracy in the dual modality setup but achieved lower scores of 28.33\,\% and 22.00\,\% with sound and touch modalities, respectively. In terms of class-specific performance on the test data, for a single class, the SVM model achieved balanced accuracies~\cite{brodersen2010balanced} of 70.68\,\% for anger, 87.65\,\% for ``Attention'', 70.68\,\% for \enquote{Happiness}, and 72.53\,\% for \enquote{Calming}. The model also performed relatively well for \enquote{Confusion} with a balanced accuracy of 66.98\,\%. However, it struggled with classes such as \enquote{Disgust} and \enquote{Fear}, showing balanced accuracies of 60.49\,\% and 59.26\,\%, respectively. 
It performs worst in \enquote{Sadness} and \enquote{Surprise} with a balanced accuracy of 54.01\,\% and 54.63\,\%, respectively.

Previous research~\cite{thompson2011effect, hertenstein2006touch} explored emotion decoding among couples and strangers, and the results we obtained in our research are similar to the results they presented for strangers (37.5\%). If we set the 37.5\% as a baseline, we used a one-sample t-test to determine whether the participant's decoding accuracy is significantly higher than chance level; the results showed that there is no significant difference between human-to-human emotion decoding among strangers in previous research and our human-robot decoding ($t(31) = -2.99, p = 0.99$).

Based on the comparison results in Table~\ref{tab:modelcomparisongestures} and Table~\ref{tab:modelcomparisonemotions}, we observed that the tactile-sound multimodal emotion recognition model achieved an accuracy of 40\,\%, demonstrating an improvement of approximately 3.89\,\% over the touch modality and 11.67\,\% over the sound modality alone. For touch gesture decoding, the tactile-sound model also showed a notable improvement, with an accuracy increase of 11.85\,\% compared to the sound modality and 9.35\,\% compared to the touch modality. These results highlight the effectiveness of integrating both tactile and auditory information, suggesting that combining the two modalities enhances the system’s ability to decode both emotions and gestures more accurately than using either modality alone.

We observed that emotion decoding based on the sound modality performed worse than the touch modality. This discrepancy could be due to the fact that participants were required to express emotions primarily through touch expressions, leading them to focus on tactile features such as amplitude, intensity, and contact area. The sound generated by touch may not fully capture the richness of these tactile expressions.

For instance, a firm grasp typically involves high force and a large contact area, which is well-detected by the tactile sensor but only produces minimal auditory feedback. Similarly, a high-speed shaking motion, which many participants used to express fear, generates significant tactile feedback but may result in weak or inconsistent sound recordings, failing to capture the full dynamics of the touch behaviour. This limitation in the sound modality could explain why participants found it more challenging to decode emotions based on auditory cues alone.

\subsubsection{Affective touch confusion matrix}
\begin{figure*}[h]
    \centering
        \includegraphics [width=\textwidth]{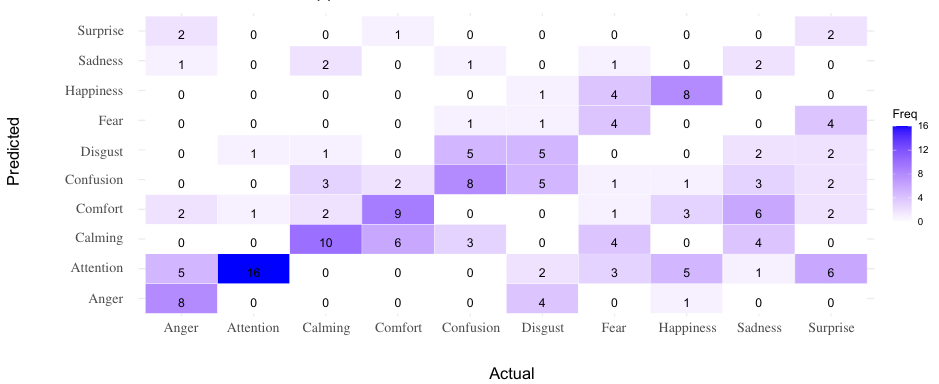}
    \caption{SVM confusion matrix for emotions.}
    \label{fig:confusion_matrix_emotion}
\end{figure*}

The confusion matrix shown in Fig.~\ref{fig:confusion_matrix_emotion} indicates that \enquote{Attention} is the most recognisable among all emotions. 
However, several misclassifications are evident: \enquote{Anger} is frequently misclassified as \enquote{Attention}, \enquote{Comfort} is often confused with \enquote{Calming}, and \enquote{Happiness} is mistakenly identified as \enquote{Attention}. 
Additionally, \enquote{Sadness} is commonly confused with both \enquote{Comfort} and \enquote{Calming}, while \enquote{Surprise} is often misclassified as \enquote{Fear} and \enquote{Attention}.

These misclassifications can be attributed to the overlapping characteristics of these emotions in the context of touch and its sound.
For example, the physical sensation and auditory feedback associated with \enquote{anger} can resemble the intense, focused signals interpreted as \enquote{Attention}. Additionally, when participants try to convey \enquote{Attention} to the robot, the maximum pressure applied is greater than that used for \enquote{Fear}, \enquote{Happiness}, and \enquote{Confusion}. This indicates that people might not be mindful of the applied force when attempting to grab the attention of a robot. \enquote{Calming} and \enquote{Comfort} are easily confused due to their similar low arousal. \enquote{Happiness}, conveyed through touch and sound, might be mistaken for the engaged and focused nature of \enquote{Attention}, and it is easily confused with \enquote{Surprise} due to their similarly high arousal. \enquote{Sadness} is often confused with similarly low arousal states of \enquote{Comfort} and \enquote{Calming}. \enquote{Surprise} and fear all belong to high arousal emotions, making them difficult to distinguish from each other.


\subsubsection{Tactile gestures confusion matrix}

\textbf{\begin{figure*}[h]
    \centering
        \includegraphics [height=6cm]{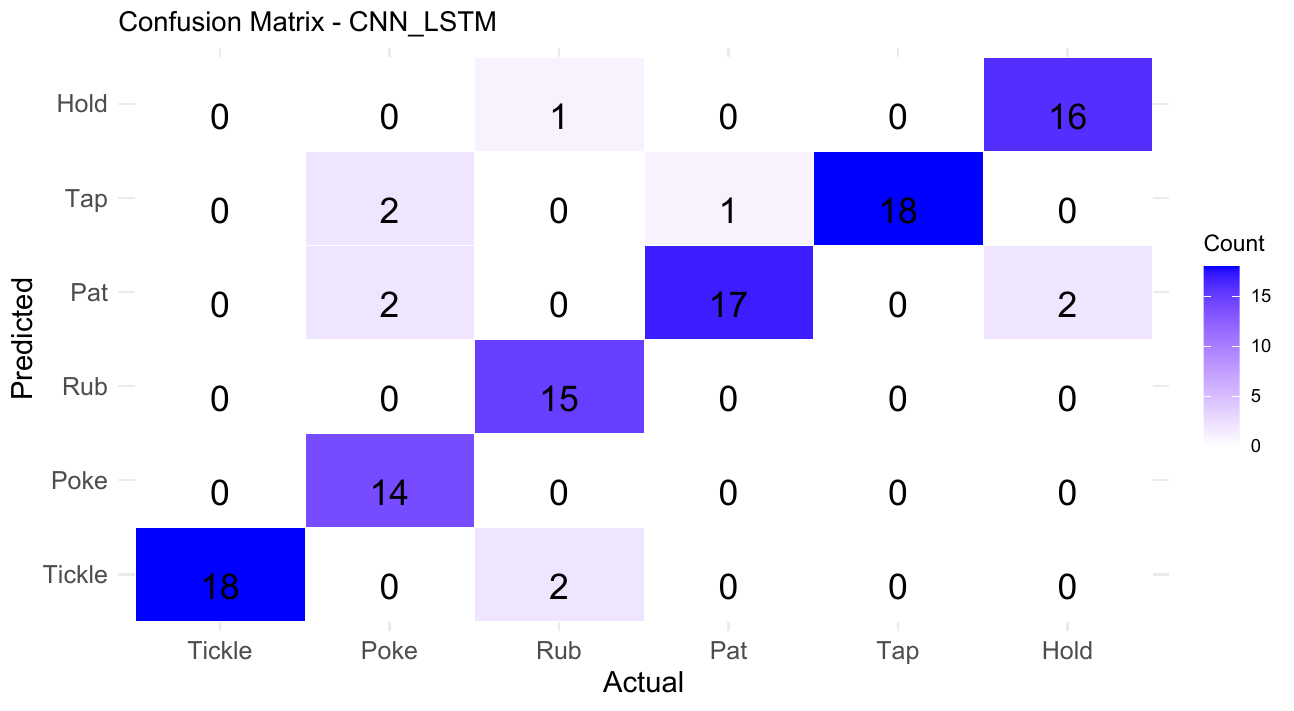}
    \caption{CNN\_LSTM confusion matrix for emotions.}
    \label{fig:confusion_matrix_gesture}
\end{figure*}
}
The overall performance of the tactile gestures surpasses that of the affective touch, which achieved 90.74\%. As shown in Fig.~\ref{fig:confusion_matrix_gesture}, the \enquote{Pat} gesture is once misclassified as \enquote{Tap}, while \enquote{Poke} is misclassified twice to \enquote{Tap} and \enquote{Pat}, respectively. In addition, \enquote{Rub} has sometimes been misclassified as \enquote{Tickle} and \enquote{Hold} and \enquote{Hold} sometimes misclassified to \enquote{Pat}, \enquote{Poke} and \enquote{Tap} are quite similar, especially when participants perform a quick poke to the robot. Additionally, \enquote{Tickle} and \enquote{Tap} are confused, which might be due to the similar motion and contact pattern involved in these gestures. \enquote{Rub}, \enquote{Hold}, and \enquote{Pat} are confused with each other, which could be due to the similar touching area of both gestures. In conclusion, the misclassification likely arises from the similar feature in execution and the overlapping sensory signal these gestures provide.


\section{Discussion}
\label{sec:discussion}


In this research, we built tactile sensing capability for the Pepper robot by designing a piezoresistive pressure sensor. One of our primary objectives was to determine whether people consistently express the same emotions and touch gestures through touch. All the emotions showed significant consistency, while the ICC indicated varying degrees of consistency across different emotions. \enquote{Attention} achieved the highest consistency, suggesting that participants were more uniform in conveying this emotion. Conversely, \enquote{Surprise} had the lowest ICC, indicating considerable variability in how participants expressed this emotion through touch. The classification accuracy of the multimodal also aligns with the ICC trend. The class \enquote{Attention} obtained the highest accuracy of 87.65\,\% while the class \enquote{Surprise} got the accuracy of 54.63\%. Similarly, all touch gestures demonstrated significant consistency, with ICC values for touch gestures generally higher than those for emotions, except for \enquote{Attention}. This suggests that people have more consensus on touch gestures than on emotions. One possible explanation is that, although emotions correlate with touch gestures, they can be expressed through various touch gestures. Additionally, people's perception of emotions can be conveyed through different touch patterns and features. In contrast, social touch gestures often have a common understanding. We also found that ``Attention'' has a higher correlation with certain gestures, such as ``Tap'' or ``Poke'', leading to higher ICC values among all emotions. This consistency in specific gestures for ``Attention'' could explain the higher ICC values and classification accuracy.

The phenomenon of varying consistency in touch gestures and emotions can be attributed to several factors. Touch gestures are often straightforward and universally understood, such as a tap or poke to convey attention, leading to higher consistency across individuals. In contrast, emotions are more complex and can be expressed through a variety of touch gestures. For example, love can be shown through a hug, a gentle touch, or a pat, resulting in lower consistency as different participants have different strategies to express different emotions. Previous research asked participants~\cite{ju2021haptic} to convey four emotions using specific gestures like tapping, rubbing, and pressing. In our research, participants were not given a specific list of gestures, leading to more variability in touch-based expression. Additionally, cultural norms and personal habits significantly influence how people express emotions through touch, with different cultures having varying interpretations and norms. Even though all participants in our study were from the same culture, personal experiences and comfort levels with touch also contributed to this variability. The context of interaction further affects how touch is interpreted, as a touch meant to convey attention in a formal setting might differ from one in a casual or intimate setting. The inherent complexity and multifaceted nature of emotions mean that a single touch gesture might not fully capture an emotion, leading to a broader range of expressions. The higher consistency in touch gestures compared to emotions is thus due to their universal nature, while the complexity and variability of emotions, influenced by cultural, personal, and contextual factors, lead to greater differences in expression.

Our findings indicate that touch gestures are generally more distinguishable than emotions when based on tactile interaction. The PERMANOVA results confirmed that all gestures were significantly different from each other, suggesting strong consensus among participants in their execution. Gestures like ``Tap'' and ``Tickle'' were among the easiest to distinguish due to their unique characteristics— ``Poke'' being a short, localized, high-intensity action and ``Rub'' involving continuous motion over a larger area. ``Tap'' was also relatively easy to decode due to its brief and repetitive nature. However, ``Pat'' and ``Tap'' were sometimes confused, likely due to similarities in their intensity and contact area. ``Pat'', in particular, shares overlapping features with both ``Tap'' and ``Tickle'', making it more prone to misclassification. In contrast, emotions were more challenging to differentiate, particularly when they shared similar tactile and auditory characteristics. Although all emotions showed statistically significant consistency, certain pairs, such as ``Calming'' and ``Sadness'' or ``Confusion'' and ``Disgust'', did not exhibit significant differences. This suggests that these emotions share common touch patterns, making them harder to distinguish. ``Happiness'' and ``Anger'' were among the easiest to decode, as they were often associated with distinct tactile expressions, ``Happiness'' frequently involved rhythmic, gentle movements, while ``Anger'' was characterized by stronger pressure and abrupt interactions. ``Fear'' and ``Disgust'' were moderately distinguishable but sometimes confused due to their overlapping aversive touch patterns.

The confusion matrix from the SVM model further revealed that emotions were often misclassified into categories with similar arousal or valence levels. For instance, emotions in the low-arousal range, such as ``Calming'' and ``Sadness'', were frequently mixed up, while high-arousal emotions like ``Surprise'' and ``Fear'' also showed overlap. Subjective feedback supported these findings, with ``Surprise'' and ``Confusion'' being the most difficult emotions to express through touch, reported by 64.3\% and 46.4\% of participants, respectively. Overall, gesture classification achieved higher accuracy than emotion classification, reinforcing the idea that social touch gestures are more consistently performed and interpreted. While some emotions can be effectively conveyed through touch, those with similar arousal or valence levels present greater challenges for clear differentiation. These findings highlight the complexity of affective touch and suggest that while gestures provide structured cues for communication, emotional expression through touch requires additional contextual or multimodal cues for enhanced recognition.


Our study confirms that multimodal models integrating touch and sound features outperform single-modal approaches in decoding emotions and gestures, achieving higher classification accuracy and robustness. The tactile-sound dual model improved emotion recognition by 3.89\% over the touch-only model and 11.67\% over the sound-only model, while also enhancing gesture decoding by 11.85\% and 9.35\% over sound and touch modalities, respectively. These improvements stem from the complementary nature of touch and sound cues, where tactile features capture intensity, amplitude, and contact area, while audio signals add temporal dynamics, compensating for the weaknesses of each modality. Single-modal sound decoding, for instance, struggled with capturing high-force gestures like firm grasps or high-speed shaking, leading to information loss. By fusing modalities, multimodal models provide richer, more diverse feature representations, making them more effective in recognizing subtle emotional and gestural variations. This has significant implications for human-robot interaction, particularly in social, assistive, and affective computing applications, where natural and adaptive emotional communication is essential. Future research should explore expanding sensor coverage, incorporating additional modalities (e.g., physiological signals), and improving context-awareness to further enhance robotic affective touch systems.

\subsection{Limitations}

However, the generalizability of this study may be limited by the relatively small sample size of 28 participants, all from a single cultural background. Additionally, the current tactile sensor is restricted to the forearm, which may not capture the full range of touch expressions used in natural interactions. Although emotions can be conveyed through touch alone, our findings show that it remains difficult to distinguish between emotions with similar arousal and valence characteristics based solely on tactile signals. In real-world scenarios, human emotional communication is inherently multimodal, involving visual cues (e.g., facial expressions, body posture) and verbal elements (e.g., tone, content of speech). The absence of these complementary modalities in our study may have limited participants’ ability to express emotions naturally and constrained the system’s decoding accuracy. Furthermore, the interpretation of touch gestures is highly context-dependent—not only influenced by the physical design of the robot but also by its perceived social role. A robot’s appearance, including its level of human-likeness and perceived warmth, can affect users’ comfort and willingness to engage in tactile interaction. For example, a more human-like and friendly-looking robot may encourage more natural and expressive touch gestures than a mechanical or utilitarian one. Similarly, the situational context—such as whether the robot is perceived as a caregiver, companion, or assistant—can shape both the types of touch used and how those touches are interpreted. These factors introduce ambiguity into the decoding process and underscore the importance of designing tactile communication systems that are sensitive to both social context and the robot’s embodied characteristics.

Future research should involve extending tactile sensor coverage to multiple body parts and conveying the emotion back to the robot. This paper highlights the potential of using tactile and auditory signals for emotion recognition in human-robot interaction, which can help the robot build an affective response during the interaction.

\section{Conclusion}
\label{sec:conclusion}

Firstly, we developed a piezoresistive sensor to enhance a robot's tactile perception, which was evidenced by the high accuracy in classifying touch gestures. Secondly, our findings demonstrated a notable consistency among participants in expressing emotions and touch gestures. Despite this, certain emotions like disgust and surprise had very low ICCs, and there were no significant differences in their features between some emotions within identical arousal or valence categories based purely on tactile expression. What's more, we observed that participants displayed greater uniformity in touch gestures than in emotional expressions, with touch gesture recognition proving to be approximately 50\% more accurate than emotion recognition. Finally, we developed two multimodal models for emotions and social gestures decoding, achieving a 40\%  accuracy rate for 10 emotions and 90.74\% accuracy for six touch gestures, which outperforms their corresponding decoding accuracy based on single modalities, including sound modality and touch modality, furthering our comprehension of how such interactions are effectively decoded.

These findings highlight the potential of tactile sensing as both a reliable channel for decoding intentional social gestures and a complementary modality for affective understanding when combined with other inputs such as sound. The significant difference between gesture and emotion decoding performance suggests that interactive systems may first benefit from recognizing structured, socially meaningful gestures before attempting to interpret more nuanced emotional states through touch alone. And emotion expression based on touch gestures is subjective, and some emotions might be limited solely based on the touch modality, which provides us with guidance for future emotion decoding, combined with other modalities to recognize those emotions better. Emotion decoding and social gestures decoding in tactile interaction are crucial in assistive contexts, such as caregiving, education, or therapy, where physical interaction plays a vital role. The observed variability in emotional expression further underscores the need for adaptive, personalized systems that can calibrate to individual or cultural touch patterns. Looking ahead, integrating tactile input with other sensory modalities in multimodal, context-aware systems could enable more natural, responsive, and socially appropriate interactions between humans and robots.



\balance

\bibliographystyle{ieeetr}
\bibliography{references}

\vfill

\end{document}